\documentclass[final]{cvpr}

\usepackage{times}
\usepackage{epsfig}
\usepackage{graphicx}
\usepackage{amsmath}
\usepackage{amssymb}

\usepackage{multirow}
\usepackage{booktabs,tabularx}
\usepackage{enumitem}
\usepackage{makecell}
\usepackage{indentfirst}
\usepackage{siunitx}
\usepackage{algorithm}
\usepackage{algorithmic}
\usepackage{subfigure}
\newcommand{\tabincell}[2]{\begin{tabular}{@{}#1@{}}#2\end{tabular}}
\usepackage[pagebackref=true,breaklinks=true,colorlinks,bookmarks=false]{hyperref}

\def\cvprPaperID{6276} 
\def\confYear{CVPR 2021}

\def\cvprPaperID{6276} 

\begin{document}

\title{Scene Text Retrieval via Joint Text Detection and Similarity Learning}
\author{
Hao Wang\textsuperscript{\rm 1},
Xiang Bai\textsuperscript{\rm 1},
Mingkun Yang\textsuperscript{\rm 1},
Shenggao Zhu\textsuperscript{\rm 2},
Jing Wang\textsuperscript{\rm 2},
Wenyu Liu\textsuperscript{\rm 1}\\
\textsuperscript{\rm 1}Huazhong University of Science and Technology,
\textsuperscript{\rm 2}Huawei Cloud \& AI\\
\{wanghao4659,xbai,yangmingkun,liuwy\}@hust.edu.cn, 
\{zhushenggao,wangjing105\}@huawei.com\\
}

\maketitle

\begin{abstract}
Scene text retrieval aims to localize and search all text instances from an image gallery, which are the same or similar to a given query text. Such a task is usually realized by matching a query text to the recognized words, outputted by an end-to-end scene text spotter. In this paper, we address this problem by directly learning a cross-modal similarity between a query text and each text instance from natural images. Specifically, we establish an end-to-end trainable network, jointly optimizing the procedures of scene text detection and cross-modal similarity learning. In this way, scene text retrieval can be simply performed by ranking the detected text instances with the learned similarity. Experiments on three benchmark datasets demonstrate our method consistently outperforms the state-of-the-art scene text spotting/retrieval approaches. In particular, the proposed framework of joint detection and similarity learning achieves significantly better performance than separated methods. Code is available at: https://github.com/lanfeng4659/STR-TDSL.
\end{abstract}

\section{Introduction}\label{sec:introduction}

In the past few years, scene text understanding has received rapidly growing interest from this community due to a large amount of everyday scene images that contain texts. Most previous approaches for scene text understanding focus on the topics of text detection, text recognition and word spotting, succeeding in many practical applications such as information security, intelligent transportation system~\cite{2018ZhuTS,RongYT16}, geo-location, and visual search~\cite{BaiYLXL18}. Different from the above topics, we study the task of scene text retrieval introduced by Mishra~\etal~\cite{MishraAJ13}, aiming to search all the text instances from a collection of natural images, which are the same or similar to a given text query. Different from a scene text reading system that must localize and recognize all words contained in scene images, scene text retrieval only looks for the text of interest, given by a user. Such a task is quite useful in many applications including product image retrieval, book search in library~\cite{YangHHOZKG17} and key frame extraction of videos~\cite{SongWHXHJ19}. 
\begin{figure}[t]
    \includegraphics[width=0.95\linewidth]{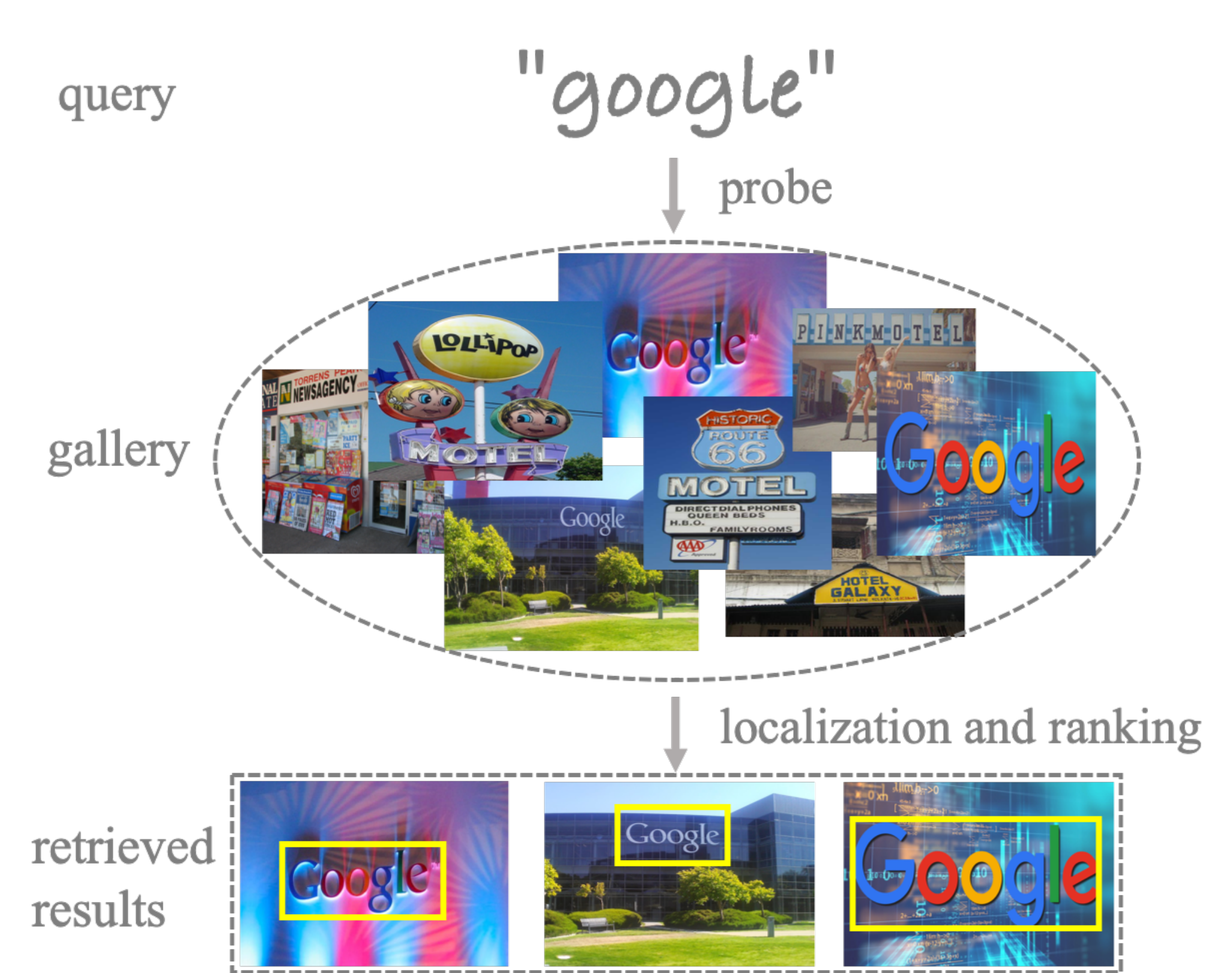}
\centering
\vspace{+0.5ex}
  \caption{Given a query text ``google'', the end-to-end scene text retrieval method aims to retrieve the images containing ``google'' from gallery, as well as their locations in the images.}
  \label{abstract}
  \vspace{-2ex}
\end{figure}

As depicted in Fig.~\ref{abstract}, the goal of scene text retrieval is to return all images that are likely to contain the query text, as well as the bounding boxes of such text. In this sense, scene text retrieval is a cross-modal retrieval/matching task that aims to close the semantic gap between a query text and each text proposal. Traditional text retrieval methods~\cite{AlmazanGFV14,AldavertRTL13,GhoshV15,Gomez-BigordaRK17} are often designed to handle cropped document text images.
Since the well-cropped bounding boxes are not easy to obtain by a detector for natural images, these methods cannot be directly applied in scene text retrieval. Mishra~\etal~\cite{MishraAJ13} first study text retrieval in scene images, which is cast as two separated sub-tasks: text detection and text image retrieval. However, the performance is limited, as their framework is designed based on handcraft features. 

Another feasible solution to scene text retrieval is based on an end-to-end text recognition system, such as~\cite{jaderberg2016reading,HeCVPR2018}. Under this setting, the retrieval results are determined according to the occurrences of the given query word within the spotted words. As verified by~\cite{MishraAJ13,GomezECCV20118YOLO+PHOC}, such methods often achieve unsatisfactory retrieval performance, as the end-to-end text recognition system is optimized with a different evaluation metric that requires high accuracy in terms of both detection and recognition. However, for a retrieval engine, more text proposals can be leveraged to reduce the performance loss brought by the missed detections. Gomez~\etal~\cite{GomezECCV20118YOLO+PHOC} directly predict the corresponding Pyramidal Histogram Of Character~\cite{AlmazanGFV14} (PHOC) of each text instance for ranking. However, these methods don't explicitly learn the cross-modal similarity.

In this paper, we propose a new deep learning framework for scene text retrieval that combines the stages of text detection and cross-modal similarity learning. Our basic idea is motivated by the recent end-to-end text spotting methods~\cite{FOTS,MaskTextspotter,ShenICCV2017,HeCVPR2018,boundary,maskv2}, which unify the two sub-tasks of text detection and text recognition with an end-to-end trainable network. Such methods usually achieve better spotting performance than those optimized separately, benefiting from feature sharing and joint optimization.
Specifically, our network for scene text retrieval is optimized by two tasks: a text detection task that aims to predict bounding boxes of candidate text instances, and a similarity learning task for measuring the cross-modal similarity between a query text and each bounding box. 
With joint optimization, the text detection task and similarity learning task are complementary to each other. On the one hand, the detection task can pay more attention to the recall than the precision in order to avoid missing detections, as the feature matching process by the similarity learning task can effectively eliminate false alarms. On the other hand, the faithful cross-modal similarity provided by the similarity learning task can be used for looking for an accurate bounding box that contains a given text. In addition, both tasks share CNN features, significantly improving the inference speed.

Unlike general objects such as person and car, scene text is a kind of sequence-like object. In order to learn a proper similarity between a query text and a text proposal, we adopt the normalized edit distances as the supervision of the pairwise loss function, which has been widely used in string matching. However, learning such similarity is not easy, as the similarity values of all word pairs for training are not evenly distributed in the similarity space. For example, the number of word pairs with low similarity is much larger than those with high similarity. As a result, the network is prone to distinguish dissimilar word pairs but difficult to similar word pairs. To ease this problem, we propose a word augmentation method: a series of pseudowords that are very similar to a given query text, are randomly generated, which are fed into the network together with the query text during the training process. 

The contribution in this work is three-fold. First, we present a new end-to-end trainable network for joint optimizing scene text detection and cross-modal similarity learning, which is able to efficiently search text instances from natural images that contain a query text. Second, we propose a word augmentation method by generating similar pseudowords as input queries, which enhances the discriminatory power of the learned similarity. Third, we collect and annotate a new dataset for Chinese scene text retrieval, consisting of 23 pre-defined query words and 1667 Chinese scene text images. This dataset is adopted to verify the effectiveness of text retrieval methods over non-Latin scripts.

\begin{figure*}[t]
    \includegraphics[width=0.98\linewidth]{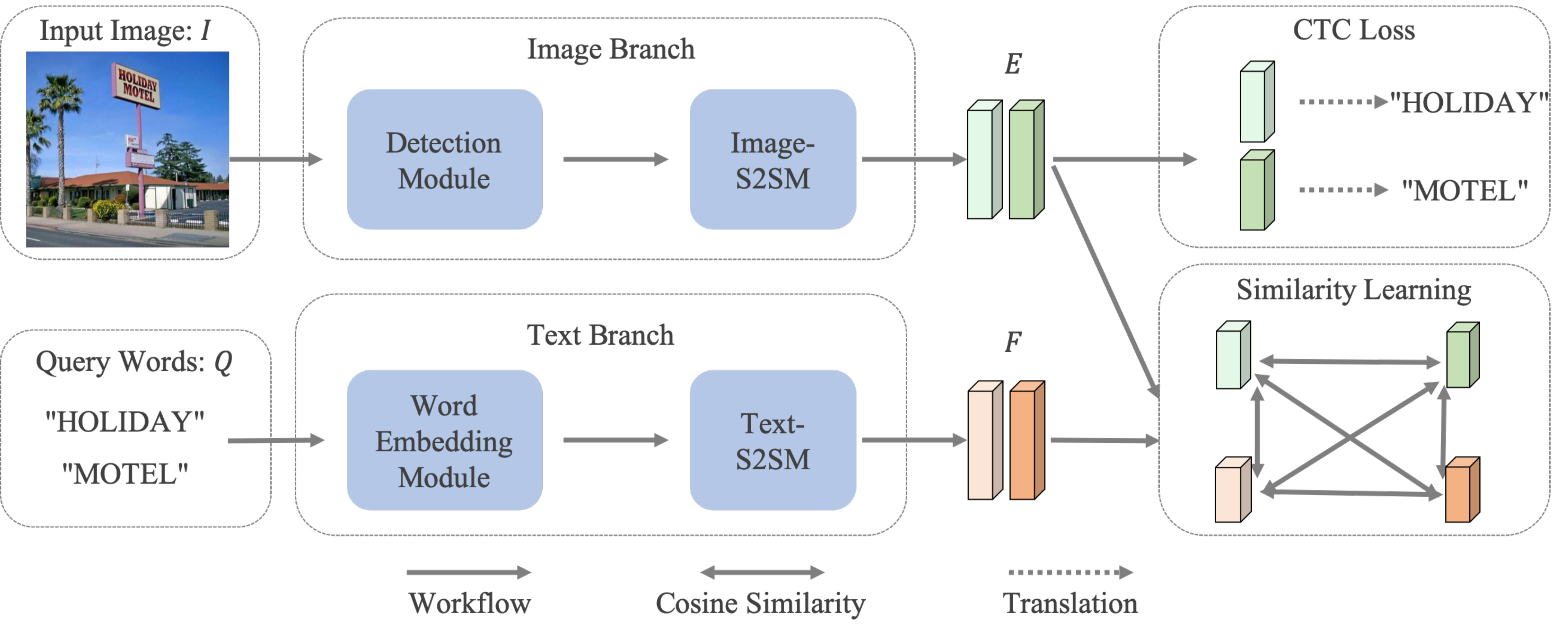}
\centering
  \caption{Illustration of our proposed framework. Given an image, text regions are detected and the corresponding features are extracted for the subsequent Image-S2SM. Meanwhile, features $F$ of all query words $Q$ are obtained through the word embedding module and the Text-S2SM. The whole network is jointly optimized by text detection task, text translation task and similarity learning task.}
\vspace{-2ex}  
  \label{framework}
\end{figure*}

\section{Related work}
Traditional text retrieval methods~\cite{AldavertRTL13,AlmazanGFV14,sudholt2016phocnet,WilkinsonB16} are proposed to retrieve cropped document text images. A popular pipeline of these methods is to represent both text string and word image and then calculate the distance between their representations. In~\cite{AlmazanGFV14}, PHOC is first proposed to represent text string for retrieval. Then, Sudholt~\etal~\cite{sudholt2016phocnet} and Wilkinson~\etal~\cite{WilkinsonB16} respectively propose to predict PHOC and DCToW from word image using neural networks. Without carefully designing a hand-crafted representation of text, Gomez~\etal~\cite{Gomez-BigordaRK17} propose to learn the Levenshtein edit distance~\cite{levenshtein1966binary} between text string and word image. The learned edit distances are directly used to rank word images. However, the above methods consider perfectly cropped word images as inputs, rather than more universal and challenging scene text images.

Mishra~\etal~\cite{MishraAJ13} first introduce the task of scene text retrieval, and adopt two separate steps for character detection and classification. Then, images are ranked by scores,~\ie, the probability of query text exists in the images. Ghosh~\etal~\cite{GhoshBKV15} propose a scene text retrieval method consisting of a selective search algorithm to generate text regions and a SVM-based classifier to predict corresponding PHOCs.
Nevertheless, these methods ignore the relevance and complementarity between text detection and text retrieval. Moreover, it is inefficient to separate scene text retrieval into two sub-tasks. To integrate text detection and text retrieval in a unified network, Gomez~\etal~\cite{GomezECCV20118YOLO+PHOC} introduce the first end-to-end trainable network for scene text retrieval, where all text instances are represented with PHOCs. Specifically, the method simultaneously predicts text proposals and corresponding PHOCs.
Then, images in the gallery are ranked according to the distance between the PHOC of a query word and the predicted PHOC of each detected text proposal.

A few text spotting methods~\cite{HeCVPR2018,jaderberg2016reading} are also adopted for scene text retrieval. These methods first detect and recognize all possible words in each scene image. Then, the probability that the image contains the given query word is represented as the occurrences of the query word within those spotted words.
Unlike retrieving text instances by matching a query word to the spotted words, our method directly measures cross-modal similarity between them.

\section{Methodology}
As illustrated in Fig.~\ref{framework}, our network consists of two branches,~\ie, image branch for extracting features $E$ of all possible text proposals and text branch that converts query words $Q = \{q_{i}\}_{i=1}^{N}$ into features $F$. Then, the pairwise similarity between $F$ and $E$ is calculated for ranking.

\subsection{Image Branch}
The image branch aims at extracting features of all possible text proposals.
Unlike general objects, scene text usually appears in the form of a character sequence. Therefore, the sequence-to-sequence module (Image-S2SM), whose structure is detailed in Tab.~\ref{S2SM}, is used to enhance the contextual information of each text proposal. As shown in Fig.~\ref{framework}, there are two modules in the image branch, including the text detection module and Image-S2SM. To simplify the detection pipeline, the text detection module is based on an anchor-free detector~\cite{tian2019fcos}, where the backbone is FPN~\cite{lin2017feature} equipped with ResNet-50~\cite{he2016deep}. 

Given an image, the detection module in the image branch first yields $K$ text proposals. The corresponding region of interest (RoI) features $P = \{p_{i}\}_{i=1}^{K}$ are extracted via RoI-Align~\cite{he2017mask} and are fed into the subsequent Image-S2SM to generate the features $E \in \mathbb{R}^{K \times T \times C}$. $T$ and $C$ stand for the width and channel of RoI features.

\subsection{Text Branch}
Different from images, query words are a set of text strings that are unable to be directly forwarded by the neural network. So a word embedding module is employed to convey query words into features. Similar to Image-S2SM, the sequence-to-sequence module (Text-S2SM), whose structure is detailed in Tab.~\ref{S2SM}, is also used in this branch.

\textbf{Word Embedding Module} consists of an embedding layer and a bilinear interpolation operator. Specifically, given query words $Q$, the word $q_{i}$ can be considered as a character sequence $(y_{1},...,y_{|q_{i}|})$, where $|q_{i}|$ is the number of characters in $q_{i}$ and $y_{j}$ is the one-hot representation of the $j$-th character of $q_{i}$. The embedding layer first converts each character $y_{j}$ into $2C$ dimensional feature, producing an embedded feature sequence for each word. Then, each feature sequence is concatenated and interpolated as a fixed-length feature $\hat{f}_{i} \in \mathbb{R}^{T \times 2C}$. Finally, all features $\{\hat{f}_{i}\}_{i=1}^{N}$ of $Q$ are stacked as output features $\hat{F} \in \mathbb{R}^{N \times T \times 2C}$.

After the word embedding module for query words, the obtained features $\hat{F}$ is projected to $F \in \mathbb{R}^{N \times T \times C}$ by Text-S2SM. Then, both features $E$ and $F$ are fed for the subsequent similarity learning task.

\subsection{Similarity learning}
After the features of text proposals and query words, $E \in \mathbb{R}^{K \times T \times C}$ and $F \in \mathbb{R}^{N \times T \times C}$, are extracted, the pairwise similarity between query words $Q$ and features of text proposals $P$ can be formulated as similarity matrix $\hat{S}(Q, P) \in \mathbb{R}^{N \times K}$. Here, the value of $\hat{S}_{i,j}(Q, P)$ is equal to the cosine similarity between features $F_{i}$ and $E_{j}$, which is formulated via
\begin{equation}
  \setlength{\abovedisplayskip}{2pt}
  \label{CosSim}
  \hat{S}_{i,j}(Q, P) = \frac{\tanh(V(F_{i}))\tanh(V(E_{j}))^{T}}{||\tanh(V(F_{i}))||*||\tanh(V(E_{j}))||}.
  \end{equation}
$V$ stands for the operator that reshapes a two-dimensional matrix into one-dimensional vector.

During training, the predicted similarity matrix $\hat{S}(Q, P)$ is supervised by the target similarity matrix $S(Q, P)$. Each target similarity, $S_{i,j}(Q, P)$ is the normalized edit distance between the corresponding word pairs ($q_{i}$, $q_{j}$), which is defined as Eq.~\ref{WordSim}. $Distance$ is the Levenshtein edit distance~\cite{levenshtein1966binary}, $|q_{i}|$ denotes the character number of $q_{i}$.
\begin{equation}
    \label{WordSim}
    S_{i,j}(Q, P) = 1-\frac{Distance(q_{i}, q_{j})}{\max(|q_{i}|, |q_{j}|)}.
\end{equation}

Besides $\hat{S}(Q, P)$, both $\hat{S}(P, P) \in \mathbb{R}^{K \times K}$ and $\hat{S}(Q, Q) \in \mathbb{R}^{N \times N}$ are also calculated for assistant training. During inference, the similarity between $q_{i}$ and an input image is equal to the maximum value of $\hat{S}_{i}(Q, P)$, which is used for ranking.

To close the semantic gap between visual feature $E$ and textual feature $F$, the connectionist temporal classification~\cite{graves2006connectionist} (CTC) loss is adopted for aligning visual feature to a text string. 
Particularly, for each $E_{i}$, a classifier consisting of a fully connected layer and a softmax layer is supervised by the associated sequence label of $q_{i}$.

 \begin{table}[t]
 \centering
 \begin{tabular}{p{.1\columnwidth}<{\centering}| p{.25\columnwidth}<{\centering}| p{.25\columnwidth}<{\centering}| p{.20\columnwidth}<{\centering}}
	\Xhline{1.0px}
 	\multirow{2}*{} & \multirow{2}*{Layers} & Parameters  & \multirow{2}*{Output Size}\\
	& & (kernel, stride) & \\
	\hline 
	\multirow{3}*{\tabincell{c}{Image-\\S2SM}} 
	& conv layer $\times$ 2 & (3,(2,1)) & ($N,2C,h,T$) \\
	 & average on $h$ & - & ($N,2C,1,T$) \\
	 & BiLSTM & - & ($N,T,C$) \\
	 \hline
	 \multirow{2}*{\tabincell{c}{Text-\\S2SM}}  
	 & conv layer & (1,(1,1)) & ($N,2C,1,T$) \\
	 & BiLSTM & - & ($N,T,C$) \\
 \Xhline{1.0px}
 \end{tabular}
 \vspace{+0.5ex}
 \caption{
  Architectures of Image-S2SM and Text-S2SM. BiLSTM stands for bidirectional LSTM \cite{hochreiter1997long} layer.
  }
  \label{S2SM}
  \vspace{-2ex}
 \end{table}

\subsection{Word augmentation strategy}
\begin{figure}[t]
\centering
\vspace{+0ex}
\includegraphics[width=0.96\linewidth]{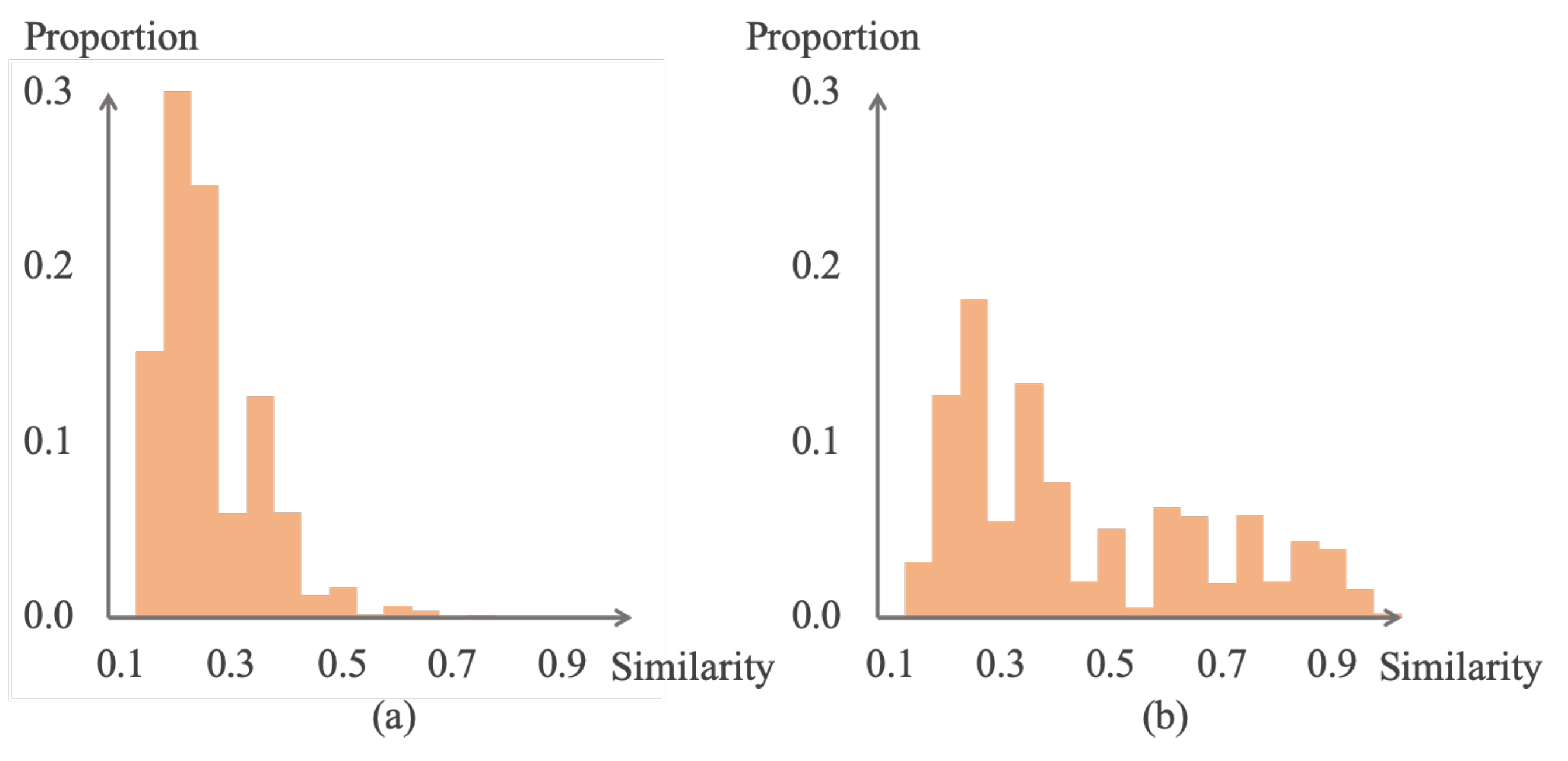}

\centering
\caption{Similarity distribution of word pairs from query words $Q$ in (a) and augmented query words $\widetilde{Q}$ in (b) on SynthText-900k.}
\label{was}
\end{figure}

We observe that the similarity learning task suffers from imbalanced distribution in the similarity space, as most random word pairs are dissimilar. As shown in Fig.~\ref{was} (a), for query words, the proportion of word pairs with low similarity is much larger than those with high similarity. As a result, the learned network tends to distinguish dissimilar word pairs but difficult to handle similar word pairs.

To alleviate this effect, a word augmentation strategy is proposed to randomly generate pseudowords that are very similar to the given query words. Then, both original query words and pseudowords are fed into the network during the training process. As illustrated in Fig.~\ref{was} (b), with such an augmentation strategy, the proportion of word pairs with high similarity is greatly enlarged, which eases the problem brought by uneven similarity distribution. 

\begin{figure}[t]
  \centering
  \vspace{+0ex}
  \includegraphics[width=0.96\linewidth]{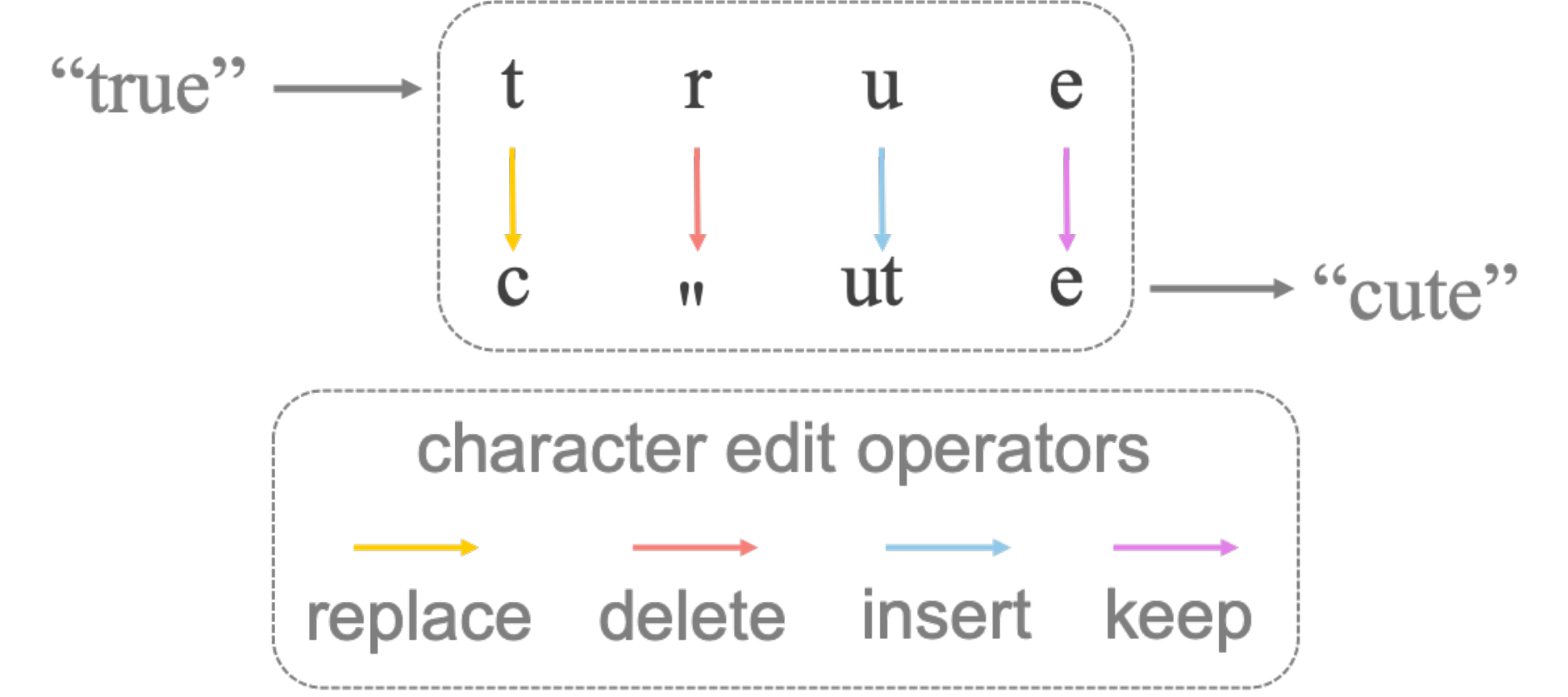}
  
  \centering
  \caption{The process of word augmentation strategy. Given an input word ``true”, this strategy outputs a word ``cute”.}
  \label{was_process}
  \vspace{-2ex}
  \end{figure}

As illustrated in Fig.~\ref{was_process}, we define four types of character edit operators,~\ie, inserting a random character behind the current character, deleting the current character, replacing the current character with a random one, and keeping the current character. For each character in the query word $q_{i}$, an edit operator is randomly chosen.
Then, a pseudoword $\hat{q}_{i}$ similar with $q_{i}$ is generated. Obviously, the higher the ratio of keeping the current character is, the more similar the pseudoword is with the input word. In our experimental setting, the sampling ratio among four character edit operators is set to 1 : 1 : 1 : 5. The algorithm description of the word augmentation strategy is summarized in Appendix A.
Finally, query words 
$\widetilde{Q} = \{\widetilde{q}_{i}\}_{i=1}^{2N}$ consisting of original words $Q$ and their associated pseudowords $\hat{Q} = \{\hat{q}_{i}\}_{i=1}^{N}$, are fed into text branch for training. In this case, the predicted pairwise similarity matrices are $\hat{S}(\widetilde{Q}, P) \in \mathbb{R}^{2N \times K}$, $\hat{S}(P, P) \in \mathbb{R}^{K \times K}$ and $\hat{S}(\widetilde{Q}, \widetilde{Q}) \in \mathbb{R}^{2N \times 2N}$.

\subsection{Loss functions}\label{subsec:loss}
The objective function consists of three parts, which is defined as follows,
\begin{equation}
    L = L_{d} + L_{s} + L_{c},
\end{equation}
where $L_{d}$ is the detection loss in \cite{tian2019fcos}.
$L_{c}$ is CTC loss for text translation task. $L_{s}$ is the cross-modal similarity learning loss, which is calculated with Smooth-L1 loss $L_{r}$ for regression. The loss function $L_{s}$ is formulated as 
\begin{equation}
\begin{aligned}
    L_{s}  
    \left.=\frac{1}{K}\sum_{i}^{K}\max(L_{r}(\hat{S}_{i}(P, P), S_{i}(P, P)))\right.\\
    \left.+\frac{1}{2N}\sum_{i}^{2N}\max(L_{r}(\hat{S}_{i}(\widetilde{Q}, P), S_{i}(\widetilde{Q}, P)))\right.\\
    \left.+\frac{1}{2N}\sum_{i}^{2N}\max(L_{r}(\hat{S}_{i}(\widetilde{Q}, \widetilde{Q}), S_{i}(\widetilde{Q}, \widetilde{Q})))\right.,
\end{aligned}
\end{equation}
where $\hat{S}$ and $S$  are the predicted similarity matrix and its associated target similarity matrix. $2N$ and $K$ are the number of query words after augmentation and the number of text instances respectively. 

\begin{table*}[t]
\centering
\begin{tabular}{m{.60\columnwidth}|m{.18\columnwidth}<{\centering} m{.18\columnwidth}<{\centering}m{.18\columnwidth}<{\centering}|m{.18\columnwidth}<{\centering}}
\Xhline{1.0pt}
Method& SVT& STR& CTR&FPS\\
\hline  
Mishra~\etal~\cite{MishraAJ13} & 56.24& 42.70& - & 0.10\\
Jaderberg~\etal~\cite{jaderberg2016reading}& 86.30& 66.50&- & 0.30\\
He~\etal~\cite{HeCVPR2018} (dictionary)& 80.54& 66.95&-& 2.35\\
He~\etal~\cite{HeCVPR2018} (PHOC)& 57.61& 46.34&- & 2.35\\
Gomez~\etal~\cite{GomezECCV20118YOLO+PHOC}& 83.74& 69.83& \ \,41.05$^{*}$& \textbf{43.50}\\
Gomez~\etal~\cite{GomezECCV20118YOLO+PHOC} (MS)& 85.18& 71.37& \ \,51.42$^{*}$& 16.10 \\
Mafla~\etal~\cite{mafla2020real}& 85.74& 71.67&- &42.20\\
ABCNet~\cite{liu2020abcnet} & \ \,82.43$^{*}$& \ \,67.25$^{*}$&- & 17.50\\
Mask TextSpotter v3~\cite{liao2020mask} &  \ \,84.54$^{*}$& \ \,74.48$^{*}$& \ \,55.54$^{*}$& 2.40 \\
\hline 
Ours & 89.38& 77.09&  66.45& 12.00\\
Ours (MS)& \textbf{91.69}& \textbf{80.16}&  \textbf{69.87}& 3.80\\
\Xhline{1.0pt}
\end{tabular}
\vspace{+1ex}
\caption{
Performance comparisons (mAP scores) with the state-of-the-art text retrieval/spotting methods on SVT, STR and CTR.
MS means multi-scale testing. $*$ represents the result is obtained with the officially released code from authors. Results with (dictionary) are filtered using a pre-defined word dictionary. Results with (PHOC) denote that recognized words are transformed to PHOC for ranking.}
\label{results_sota}
\vspace{-2ex}
\end{table*}

\section{Experiments}
First, we introduce the datasets used in our experiments and the new dataset created by us. Then, the implementation details are given. Third, we evaluate our method and make comparisons with the state-of-the-art scene text retrieval/spotting approaches. Last, we provide the ablation studies and validate a potential application of our method.

\subsection{Datasets}
 \textbf{Street View Text} dataset (SVT)~\cite{svt} has 349 images collected from Google Street View. This dataset is divided into a train set of 100 images and a test set of 249 images. 427 annotated words on the test set are used as query text.

\textbf{IIIT Scene Text Retrieval} dataset (STR)~\cite{MishraAJ13} consists of 50 query words and 10,000 images. It is a challenging dataset due to the variation of fonts, styles and view points.

\textbf{Coco-Text Retrieval} dataset (CTR) is a subset of Coco-Text~\cite{COCOText}. We select 500 annotated words from Coco-Text as queries. Then, 7196 images in Coco-Text containing such query words are used to form this dataset.       

\textbf{SynthText-900k} dataset~\cite{GomezECCV20118YOLO+PHOC} is composed of about 925,000 synthetic text images, generated by a synthetic engine~\cite{GuptaVZ16} with slight modifications.

\textbf{Multi-lingual Scene Text 5k} dataset (MLT-5k) is a subset of MLT~\cite{MLT19}, which consists of about 5000 images containing text in English.

All text instances from these datasets are in English. In our experiments, SVT, STR and CTR are the testing datasets, while SynthText-900k and MLT-5k are only used for training the proposed model.  

\begin{figure}[t]
    \includegraphics[width=0.9\linewidth]{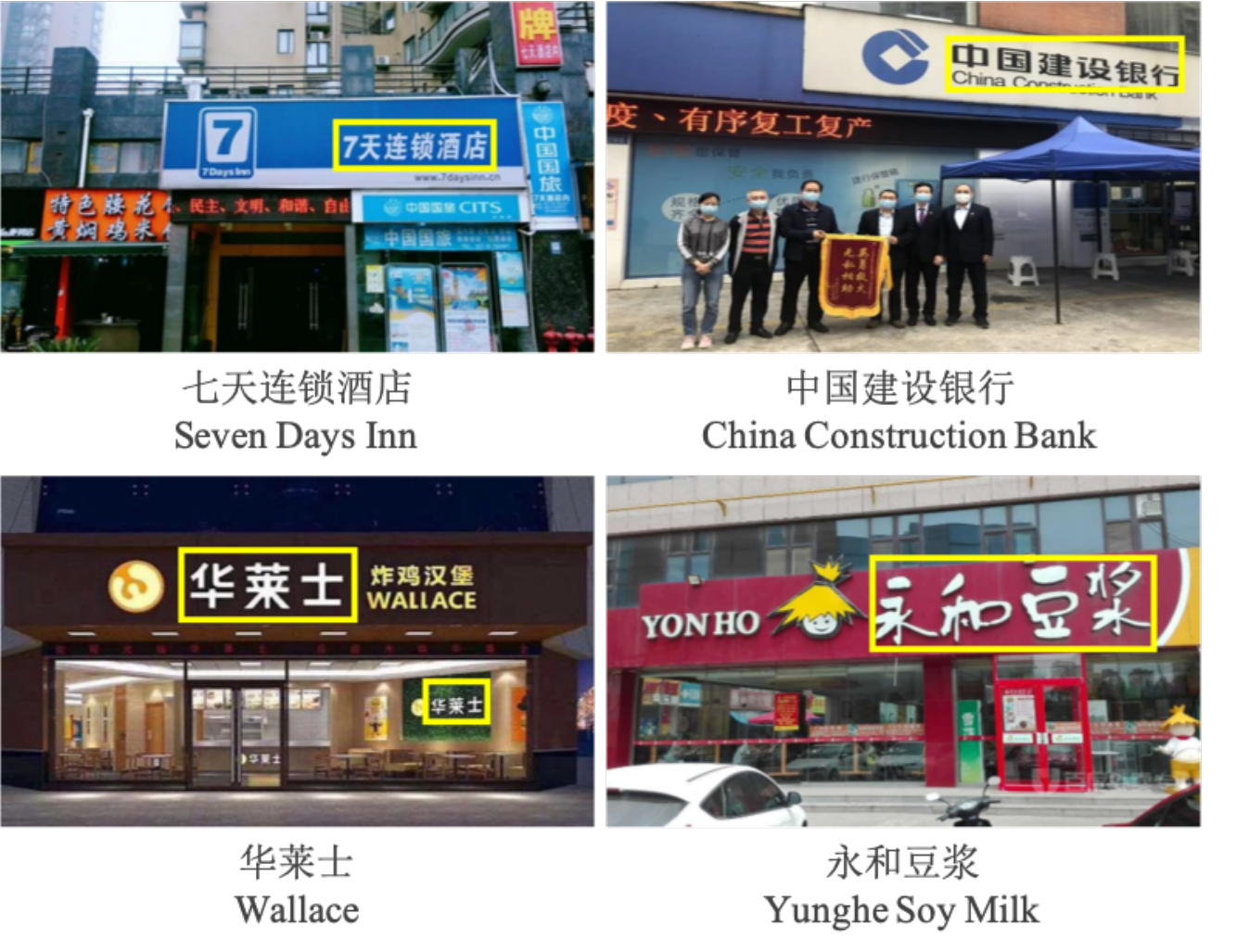}
\centering
  \caption{Examples from our CSVTR dataset.}
  \label{example_query}
  \vspace{-2ex}
\end{figure}

In order to validate our method's effectiveness on non-Latin scripts, we create a new dataset for Chinese scene text retrieval, named as~\textbf{Chinese Street View Text Retrieval} dataset (CSVTR\footnote{https://github.com/lanfeng4659/STR-TDSL}). This dataset consists of 23 pre-defined query words in Chinese and 1667 Chinese scene text images collected from the Google image search engine. Each image is annotated with its corresponding query word among the 23 pre-defined Chinese query words.
As shown in Fig.~\ref{example_query}, most query words are the names of business places, such as~\textit{Seven Days Inn},~\textit{China Construction Bank},~\textit{Wallace}, and~\textit{Yunghe Soy Milk}.    

\subsection{Implementation details}
The whole training process includes two steps: pre-trained on SynthText-900k and fine-tuned on MLT-5k. In the pre-training stage, we set the mini-batch to 64, and input images are resized to 640 $\times$ 640. In the fine-tuning stage, data augmentation is applied. Specifically, the longer sides of images are randomly resized from 640 pixels to 1920 pixels. Next, images are randomly rotated in a certain angle range of [$-15^{\circ}, 15^{\circ}$]. Then, the heights of images are rescaled with a ratio from 0.8 to 1.2 while keeping their widths unchanged. Finally, 640 $\times$ 640 image patches are randomly cropped from the images. The mini-batch of images is kept to 16.

We optimize our model using SGD with a weight decay of 0.0001 and a momentum of 0.9. In the pre-training stage, we train our model for 90k iterations, with an initial learning rate of 0.01. Then the learning rate is decayed to a tenth at the 30k$^{th}$ iteration and the 60k$^{th}$ iteration, respectively. In the fine-tuning stage, the initial learning rate is set to 0.001, and then is decreased to 0.0001 at the 40k$^{th}$ iteration. The fine-tuning process is terminated at the 80k$^{th}$ iteration. For testing, the longer sides of input images are resized to 1280 while the aspect ratios are kept unchanged. 

All experiments are conducted on a regular workstation with NVIDIA Titan Xp GPUs with Pytorch. The model is trained in parallel on four GPUs and tested on a single GPU.

\subsection{Experimental results} \label{exp_on_ben}

The experimental results of previous state-of-the-art methods and our method are summarized in Tab.~\ref{results_sota}.
Note that two most recent state-of-the-art scene text spotting approaches, namely Mask TextSpotter v3~\cite{liao2020mask} and ABCNet~\cite{liu2020abcnet} are also adopted in the comparisons. Specifically, edit distances between a query word and the spotted words from scene images are used for text-based image retrieval. The retrieval results by Mask TextSpotter v3 and ABCNet are obtained with their officially released models\footnote{https://github.com/MhLiao/MaskTextSpotterV3}$^,$\footnote{https://github.com/Yuliang-Liu/bezier\_curve\_text\_spotting}.

\subsubsection{Comparisons with state-of-the-art methods}
We can observe that the proposed method outperforms other methods by a significant margin over all datasets. 
The method~\cite{jaderberg2016reading} exhibits the best performance among previous methods on SVT, but still performs worse than our method by 3.08\%.
Compared with other methods, Mask TextSpotter v3 performs best on STR and CTR, yet our method still obtains 2.61\% and 10.91\% improvements. Moreover, our method runs much faster than Mask Textspotter v3 with single scale testing. Due to the equipment with a fast detector, YOLO~\cite{redmon2017yolo9000}, these PHOC-based methods~\cite{GomezECCV20118YOLO+PHOC,mafla2020real} achieve the fastest inference speed. However, our method obtains significantly superior performance than PHOC-based methods among all datasets. The consistent improvements over both PHOC-based methods and end-to-end text spotters further demonstrate that it is essential to learn a cross-modal similarity for scene text retrieval.

To further boost the performance, we use multi-scale testing by combing the outputs under multiple resolutions of test images. In this experiment, longer sides of input images are resized to 960, 1280 and 1600. Compared with single scale testing, the performance is improved by 2.31\%, 3.07\% and 3.42\% on SVT, STR and CTR.

\begin{figure*}[t]
\centering
\centering
\includegraphics[width=0.9\linewidth]{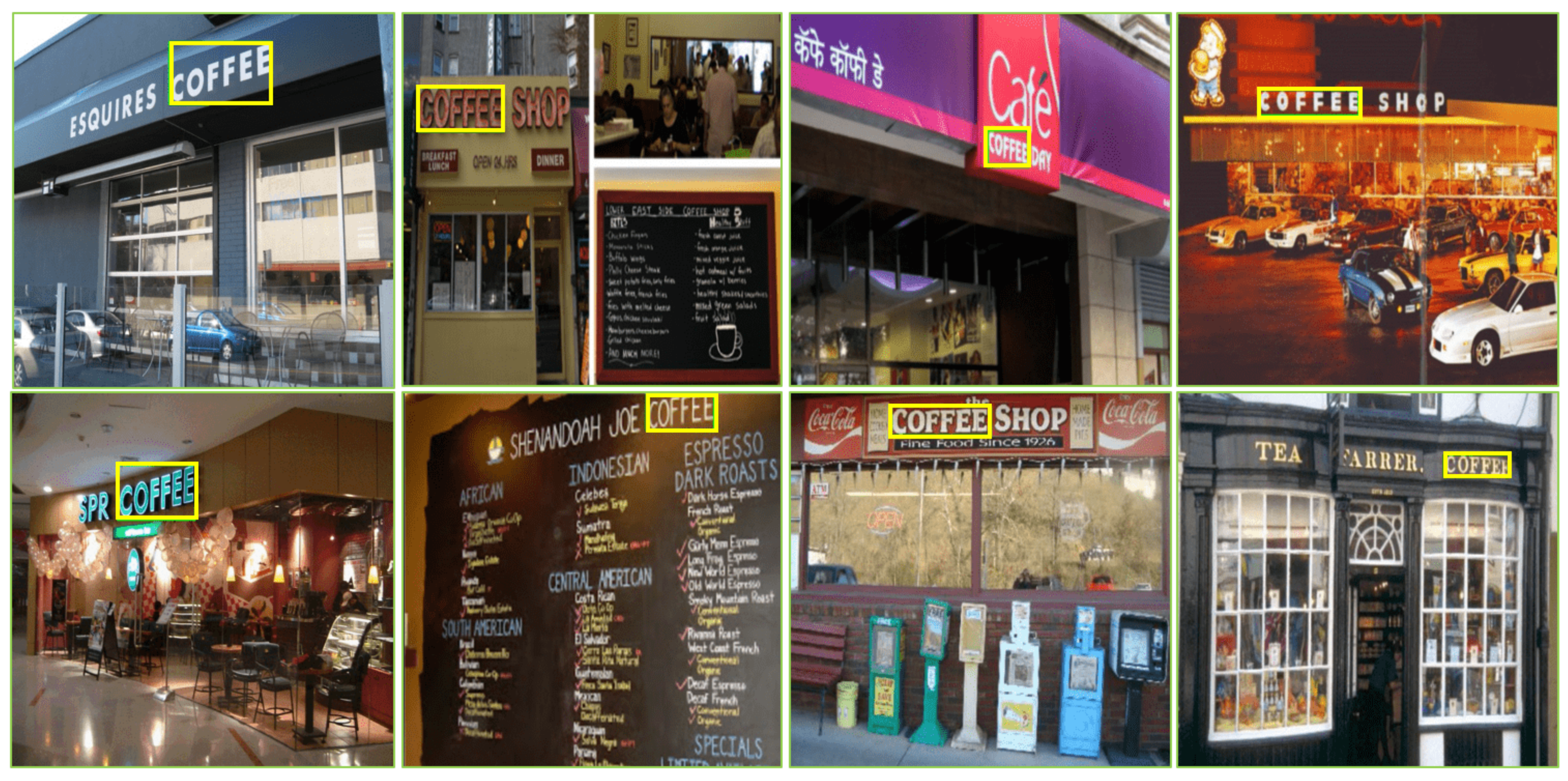}
\centering
\caption{Retrieval results on STR. Only top 8 retrieved images are shown for the query word ``coffee”.
}
\label{visualization}
\end{figure*}

Some qualitative results of our method are shown in Fig.~\ref{visualization}. We can find that the proposed method can accurately localize and retrieve text instances from an image gallery with the given query word.

\subsubsection{Chinese scene text retrieval}
To further confirm the generality of our method over non-Latin scripts, we conduct experiments on the CSVTR dataset for Chinese scene text retrieval. Our method is compared with the recent end-to-end scene text retrieval method~\cite{mafla2020real} with its officially released code\footnote{https://github.com/AndresPMD/Pytorch-yolo-phoc}. Following~\cite{GuptaVZ16}, we first generate 70k synthesized images with horizontal text instances in Chinese and their corresponding annotations. Then, the models of our method and~\cite{mafla2020real} are trained with this synthesized dataset and evaluated on the CSVTR dataset.

According to the definition of PHOC~\cite{AlmazanGFV14}, 
the dimension of PHOC dramatically increases as the number of character types grows.
Due to the limitation of GPU memory, only around 1500 Chinese character types are allowed for PHOC in~\cite{mafla2020real} during training. In order to successfully apply the PHOC-based retrieval method to Chinese scene text retrieval, we combine 1000 most common Chinese characters and characters in CSVTR to construct the character set of 1019 categories. In Tab.~\ref{csvtr_result}, our method achieves 60.23\% mAP under the setting of 1019 character types. However, the method in~\cite{mafla2020real} only performs 4.79\% mAP, and its inference speed decreases to just 1.7 FPS, while our method remains 12 FPS. 
Moreover, our method could support all 3755 character categories appearing in the training dataset and achieve 50.12\% mAP. These results reveal that our method is more robust and can be easily generalized to non-Latin scripts. Some qualitative results of retrieving Chinese text instances are shown in Appendix B.

\begin{table}[t]
  \centering
  \begin{tabular}{m{.27\columnwidth}|m{.13\columnwidth}<{\centering}|m{.13\columnwidth}<{\centering}|m{.1\columnwidth}<{\centering}|m{.1\columnwidth}<{\centering}}
  \Xhline{1.0pt}
  Method& Fea. Dim.& Char. Num.& mAP& FPS\\
  \hline
  Mafla~\etal~\cite{mafla2020real}& 14266& 1019& 4.79& 1.70\\
  Ours& 1920& 1019& 60.23& 12.00\\
  Ours$^\dagger$ & 1920& 3755& 50.12& 12.00\\
  \Xhline{1.0pt}
  \end{tabular}
  \vspace{+0.8ex}
  \caption{Feature dimension, retrieval performance and inference time comparisons on CSVTR dataset. \textit{Fea. Dim.} and \textit{Char. Num.} denote feature dimension and the number of character types.}
  \label{csvtr_result}
  \vspace{-2ex}
  \end{table}

\subsection{Ablation study}

In this section, we first provide elaborate ablation studies to further verify the effectiveness of the proposed word augmentation strategy (WAS) and CTC loss. The results are reported in Tab.~\ref{ablation_ctc_was_e2e}. The \textit{Baseline} is the basic method based on similarity learning without WAS and CTC loss.
Then we discuss the influences of assistant training by S(P, P) and S(Q, Q), and compare our cross-modal similarity learning method with the PHOC-based method on the same detection network in Tab.~\ref{ablation_fcos}.

\textbf{WAS.} As we discussed before, WAS is important for cross-modal similarity learning. Compared with the \textit{Baseline} method, WAS respectively increases the mAP by 0.63\%, 2.06\% and 1.62\% on SVT, STR and CTR datasets. The results reveal the importance of WAS, as it can enlarge the proportion of similar word pairs to prevent the model from suffering from imbalanced sample distribution in the training stage. Such an augmentation strategy facilitates the retrieval model to better distinguish similar words.
\begin{table}[t]
  \centering
  \begin{tabular}{m{.32\columnwidth}|m{.1\columnwidth}<{\centering}m{.1\columnwidth}<{\centering}m{.1\columnwidth}<{\centering}}
  \Xhline{1.0pt}
  Method& SVT& STR& CTR\\
  \hline
  Baseline& 88.05& 73.70& 62.97\\
  + CTC& 88.62& 75.54& 64.08\\
  + WAS& 88.68& 75.76& 64.59\\
  + WAS + CTC& 89.38& 77.09& 66.45\\
  \hline
  Separated& 82.88& 70.83& 61.16\\
  \Xhline{1.0pt}
  \end{tabular}
  \vspace{+1ex}
  \caption{The ablation studies on WAS and CTC. ``+ \textit{Component}'' denotes adding ``\textit{Component}'' to the Baseline method. Comparisons with the method learning in separated manner are also given. The numbers in the table stand for mAP scores.}
  \label{ablation_ctc_was_e2e}
  \vspace{-2ex}
  \end{table}
  
  \begin{table}[t]
    \vspace{2ex}
  \centering
  \begin{tabular}{m{.32\columnwidth}|m{.1\columnwidth}<{\centering}m{.1\columnwidth}<{\centering}m{.1\columnwidth}<{\centering}}
  \Xhline{1.0pt}
  Method& SVT& STR& CTR\\
  \hline
  Detection+PHOC& 80.87& 60.00& 42.89\\
  Ours$^\ddagger$& 85.28& 66.50& 51.79\\
  Ours& 87.78& 70.18& 53.90\\
  \Xhline{1.0pt}
  \end{tabular}
  \vspace{+1ex}
  \caption{Ablation study with PHOC. All models are trained with SynthText-900k. $^\ddagger$ denotes training without S(P, P) and S(Q, Q). The numbers in the table stand for mAP scores. }
  \label{ablation_fcos}
  \vspace{-2.0ex}
  \end{table}

\begin{figure}[t]
    \includegraphics[width=0.92\linewidth]{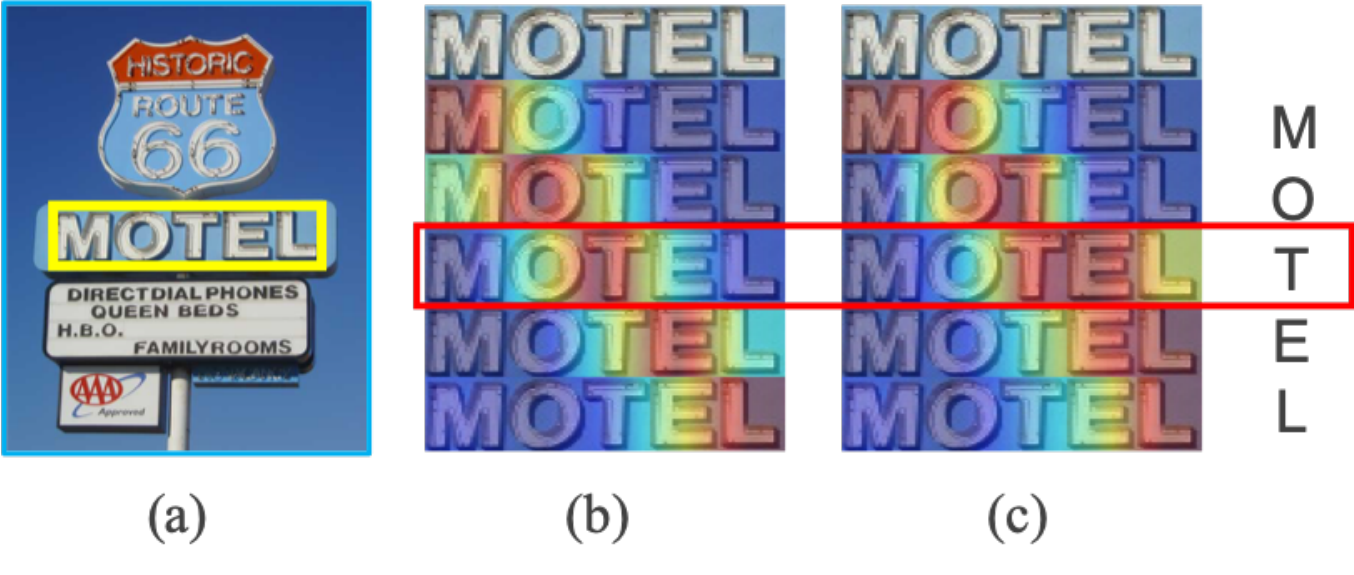}
\centering
  \caption{(a): The retrieved image with the query word “MOTEL”. The visualization of similarity between each character of “MOTEL” with text image when the network is trained with CTC loss (b) or without CTC loss (c).}
  \label{ctc_impact}
  \vspace{-2ex}
\end{figure}

\textbf{CTC loss.} Learning cross-modal similarity is of great importance in our method. CTC loss is adopted to align the cross-modal features of query words and text instances. From Tab.~\ref{ablation_ctc_was_e2e}, we can observe that CTC loss also obtains improvements of 0.57\%, 1.84\% and 1.11\% on SVT, STR and CTR, compared with the~\textit{Baseline} method. As shown in Fig.~\ref{ctc_impact}, such a process assists the image branch to focus on text regions and thus extract purified visual features, which facilitates the cross-modal similarity measure.
Therefore, similarity learning can benefit from aligning the cross-modal features via the text recognition task.

After combining the two components, the performance is further improved on all datasets, which demonstrates WAS and CTC loss are complementary for text retrieval.

\textbf{Ours vs.~Separated.} In general, scene text retrieval methods in an end-to-end manner could achieve better performance than methods with separated detection and retrieval, due to feature sharing and joint optimization.
To confirm this assumption, we separate our scene text retrieval pipeline into scene text detection and text image retrieval. The text detection network is first trained. Then, text images are cropped according to annotated bounding boxes to train the scene text retrieval network. For a fair comparison, the scene text retrieval network consists of a ResNet-50 backbone and a similarity learning network. And the similarity learning network is composed of a word embedding module, Image-S2SM and Text-S2SM. During inference, text images are cropped according to the detection results from the text detection network, and then fed to the text image retrieval network for ranking. The results are shown in Tab.~\ref{ablation_ctc_was_e2e}. Compared with the separated retrieval system, the end-to-end retrieval system can respectively improve the performance by 6.50\%, 6.26\% and 5.29\% on SVT, STR and CTR, demonstrating its effectiveness.

\textbf{Ours vs.~PHOC.} For a fair comparison with PHOC-based methods~\cite{GomezECCV20118YOLO+PHOC,mafla2020real} on the same text detector, we drop out the text branch and add a classifier behind Image-S2SM to predict the PHOC vector. The better results achieved in Tab.~\ref{ablation_fcos} demonstrate the effectiveness of the cross-modal similarity method. The reason behind might be PHOC embedding does not explicitly learn a faithful similarity between a word and a word image.

\textbf{The impact of S(P, P) and S(Q, Q).} Both S(P, P) and S(Q, Q) are only utilized for training. As shown in Tab.~\ref{ablation_fcos}, the performance degrades 2.5\% mAP on average among all datasets if they are removed.

\subsection{Application to weakly supervised text annotation}

In this section, we extend our retrieval method to a novel application. 
Nowadays, there are a large number of images with corresponding descriptions on the Internet.
Given the associated words, our method is able to precisely localize them in images, which is essentially a weakly supervised text annotation task. 
Specifically, for each word appearing in the image, its region annotation can be formulated as the bounding box of text proposal that is most similar to the word.
We evaluate the annotation quality using the scene text detection protocol in~\cite{IC13}.

\begin{figure}[t]
    \includegraphics[width=0.98\linewidth]{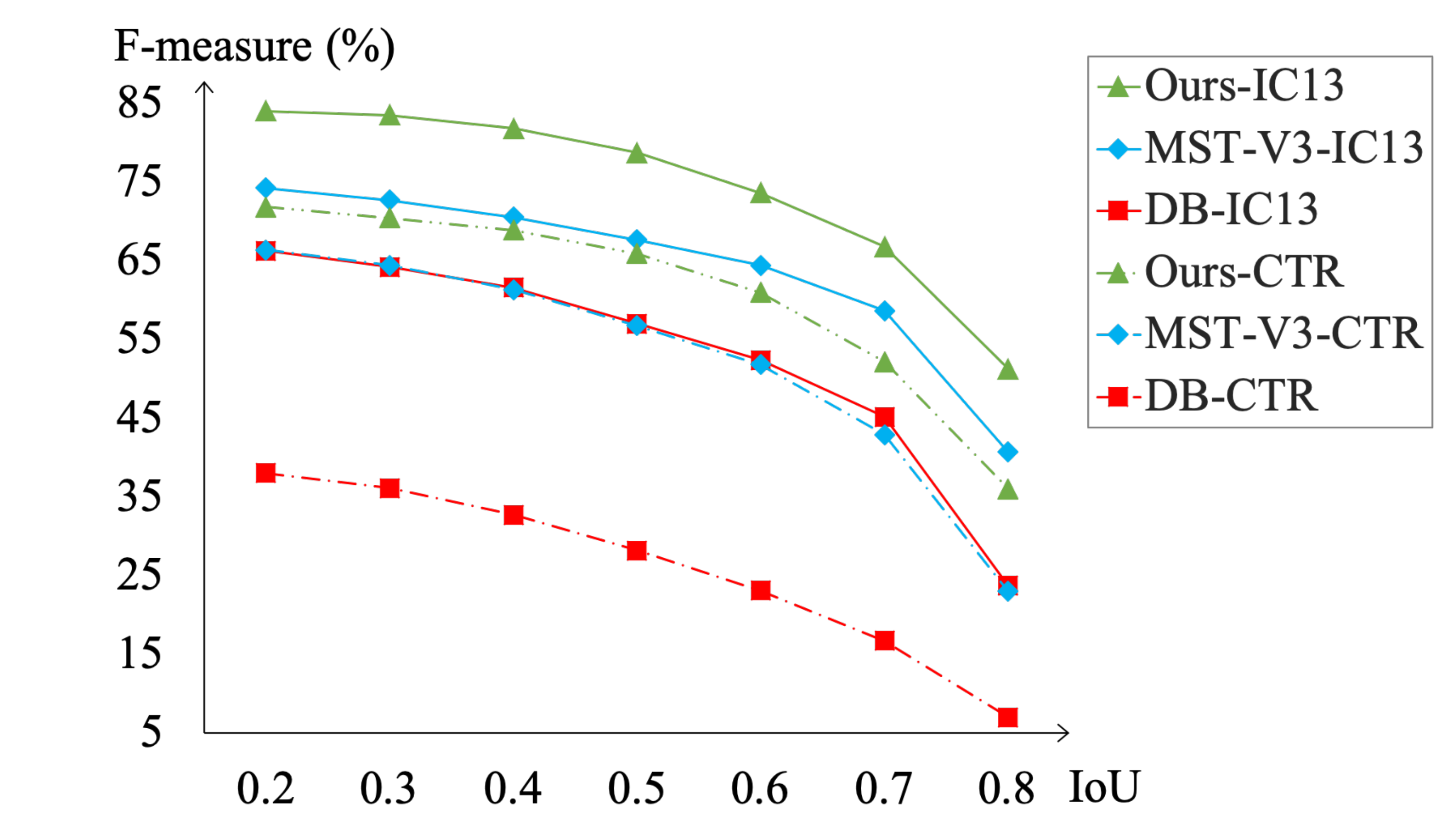}
\centering
  \caption{Comparison of text detection results on IC13 and CTR. MST-V3 denotes Mask TextSpotter v3.}
  \label{weak_anno}
  \vspace{-2ex}
\end{figure}

\begin{figure}[t]
    \includegraphics[width=0.98\linewidth]{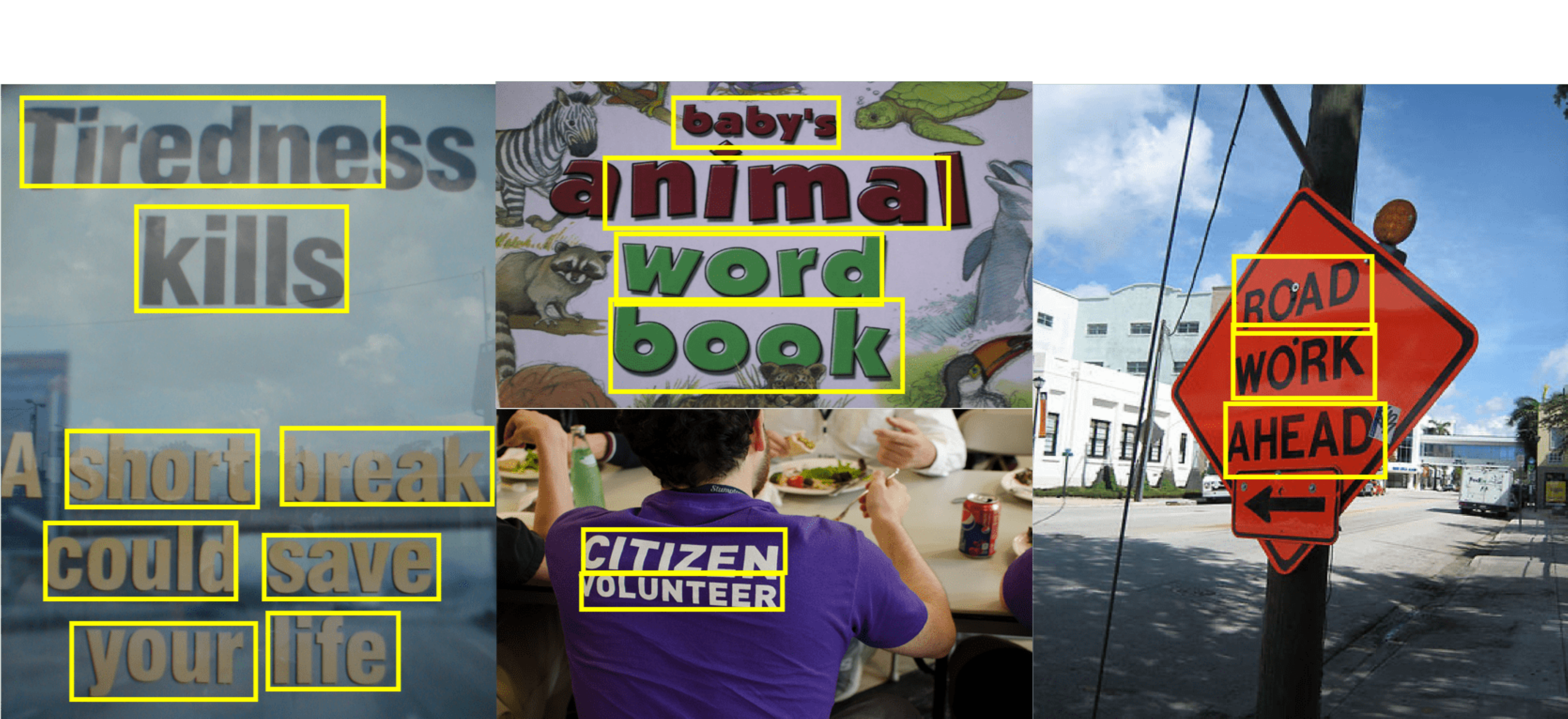}
\centering
  \caption{The bounding boxes annotated by our method on IC13 and CTR.}
  \label{weak_examples}
  \vspace{-2ex}
\end{figure}

To demonstrate its effectiveness, a recent text detector DB~\cite{liao2020real} and an end-to-end text spotter Mask TextSpotter v3~\cite{liao2020mask} are adopted for comparisons. 
Firstly, the models of DB, Mask TextSpotter v3 and our proposed method are consistently trained with the SynthText-900k. 
Then, the detected results by DB, the spotted text boxes by Mask TextSpotter v3, and the retrieved text proposals by our method are used for evaluation on IC13~\cite{IC13} and CTR. 
As shown in Fig.~\ref{weak_anno}, our method outperforms Mask TextSpotter v3 with improvements of over 10.0\%. Furthermore, compared with the text detector DB, our method achieves significant improvements of over 20.0\% and 35.0\% on IC13 and CTR. Some qualitative results of the annotated text instances on IC13 and CTR are shown in Fig.~\ref{weak_examples}. We can observe that the bounding boxes of most text instances are annotated accurately.

\section{Conclusion}
In this paper, we have presented an end-to-end trainable framework combining scene text detection and pairwise similarity learning, which could search the text instances that are the same or similar with a given query text from natural images.
The experiments demonstrate that the proposed method consistently outperforms the state-of-the-art retrieval/spotting methods on three benchmark datasets. Besides, it has been shown that our framework could cope with Chinese scene text retrieval, which is more challenging for existing methods. In the future, we would like to extend this method in dealing with more complicated cases, such as multi-oriented or curve text.


{\small
\bibliographystyle{ieee_fullname}
\bibliography{egbib}
}

\clearpage

\centerline{\textbf{\Large{Appendix}}}
\section*{A.\quad The details of word augmentation strategy}\label{sulwas}

The algorithm below describes the details of the word augmentation strategy. Given an input word $M$, this algorithm outputs a word $W$.
$MD_{4}$ stands for multinomial distribution on the variables that indicate the four types of character edit operators. And $p_{i}$ is the probability of sampling the $i$th character edit operator. In our experimental setting, the sampling ratio among four character edit operators is set to 1 : 1 : 1 : 5. $l$ is the number of character types. $n$ is the character number of the input word $M$. $j$ is a random number in the range $[0, l-1]$.
\begin{algorithm}
    \caption{Word Augmentation Strategy}
    \label{alg:A}
    \begin{algorithmic}[1]
        \REQUIRE ~~ \\
        Character set: $B = \{b_0, b_1,..., b_{l-1} \}$.\\
        Multinomial distribution: $MD_{4}(4: p_{1},...,p_{4})$.\\
        \textbf{Input:} Word $M = [m_0, m_1,..., m_{n-1}], m_{i} \in B$.\\
        \ENSURE ~~ \\
        \textbf{Output:} Word $W = []$.\\
        \STATE Generate random sequence $A = (a_{0},...,a_{n-1})$,\\ $A \sim MD_{4}(4: p_{1},...,p_{4})$,\\ $a_{i} \in \{insert, delete, replace, keep \}$.
        \FOR{i=0; i $<$ n; i++}
            \IF{$a_{i} = insert$} 
            \STATE W.append($m_{i}$)
            \STATE W.append($b_{j}$)
            \ENDIF
            \IF{$a_{i} = delete$}
            \STATE continue
            \ENDIF
            
            \IF{$a_{i} = replace$}
            \STATE W.append($b_{j}$)
            \ENDIF
            
            \IF{$a_{i} = keep$}
            \STATE W.append($m_{i}$)
            \ENDIF
        \ENDFOR
    \end{algorithmic}
\end{algorithm}
\newpage
\section*{B.\quad Examples of Chinese scene text retrieval results}\label{sulctr}
\begin{figure}[h]
\centering

\includegraphics[width=0.85\linewidth]{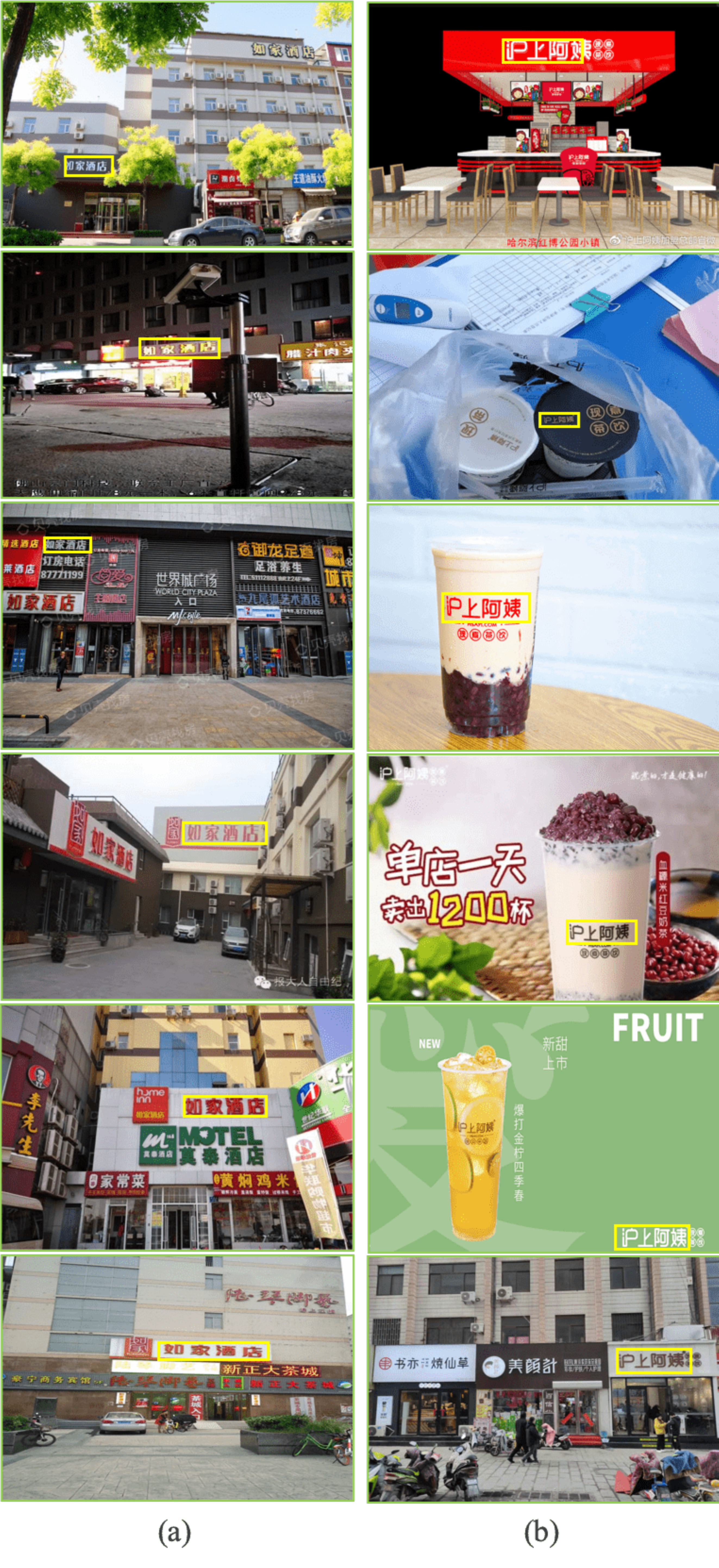}
\centering
\caption{Retrieval results on CSVTR. Only top 6 retrieved images are shown for query words ``Home Inn'' in (a) and ``Auntea Jenny'' in (b).}
\label{chinese_retrieval}
\end{figure}

Some qualitative results of Chinese scene text retrieval are shown in Fig.~\ref{chinese_retrieval}. We can observe that our method not only obtains the correct ranking lists but also localizes the bounding boxes of text instances.


 
\end{document}